# Malicious Web Domain Identification using Online Credibility and Performance Data by Considering the Class Imbalance Issue

Zhongyi Hu[a], Raymond Chiong[b,*], Ilung Pranata[b], Yukun Bao[c], and Yuqing Lin[b]

[a] School of Information Management, Wuhan University
Wuhan 430072, China
[b] School of Electrical Engineering and Computing, The University of Newcastle
Callaghan, NSW 2308, Australia
[c] School of Management, Huazhong University of Science and Technology
Wuhan, 430074, China

**Abstract**
**Purpose –** Malicious web domain identification is of significant importance to the security protection of Internet users. With online credibility and performance data, this paper aims to investigate the use of machine learning techniques for malicious web domain identification by considering the class imbalance issue (i.e., there are more benign web domains than malicious ones).
**Design/methodology/approach –** We propose an integrated resampling approach to handle class imbalance by combining the Synthetic Minority Over-sampling TEchnique (SMOTE) and Particle Swarm Optimisation (PSO), a population-based meta-heuristic algorithm. We use the SMOTE for over-sampling and PSO for under-sampling.
**Findings –** By applying eight well-known machine learning classifiers, the proposed integrated resampling approach is comprehensively examined using several imbalanced web domain datasets with different imbalance ratios. Compared to five other well-known resampling approaches, experimental results confirm that the proposed approach is highly effective.
**Practical implications –** This study not only inspires the practical use of online credibility and performance data for identifying malicious web domains, but also provides an effective resampling approach for handling the class imbalance issue in the area of malicious web domain identification.
**Originality/value –** Online credibility and performance data is applied to build malicious web domain identification models using machine learning techniques. An integrated resampling approach is proposed to address the class imbalance issue. The performance of the proposed approach is confirmed based on real-world datasets with different imbalance ratios.
**Paper type** Research paper

## 1 Introduction

Online malicious attacks represent a big threat to Internet users' privacy and security (Dong-Her et al. 2004). Phishing attacks, which deceive users into sharing passwords and their private information, and malware attacks, which secretly access and infect users' computers by distributing viruses and malicious software, are two main types of malicious attacks. These attacks not only result in immediate monetary loss, but also shatter users' trust of engaging in future online activities (San-Martín and Jimenez 2017).

In light of the above concerns, numerous studies have investigated different approaches to identify malicious web domains. These approaches can be categorised into the following two types: *blacklist*-based and *machine learning*-based approaches. The blacklist-based approach, which is a typical way to identify malicious sites, detects malicious web domains by comparing the domains a user visits against a user-verified blacklist (https://crypto.stanford.edu/SpoofGuard/). Although it is very straightforward and has been widely used in different browser toolbars, some studies have confirmed the ineffectiveness of such techniques (e.g., Tsai et al. (2011) and Purkait (2015)). The main reason being it is not only challenging but almost impractical for any blacklist to be up-to-date all the time (Ma et al. 2009; Pranata *et al.* 2012). A more effective way to tackle the problem is to identify malicious web domains using machine learning techniques. Many popular machine learning techniques have been successfully applied to identify malicious websites with lexical and host-based features extracted from URLs (Ma et al. 2009; Ma et al. 2009; Blum et al. 2010; Abutair and Belghith 2017).

---

* Corresponding author
 *Email address*: Raymond.Chiong@newcastle.edu.au (Raymond Chiong)

Some studies also built malicious web domain identification models using machine learning techniques with features extracted from web page content (Zhang et al. 2011; Moghimi and Varjani 2016; Tan et al. 2016). Instead of extracting features from URLs or web page content, we explored the use of online credibility and performance data to identify malicious web domains with machine learning techniques (Hu et al. 2016). Our results confirmed that with such kind of data, the examined machine learning approaches could accurately identify malicious web domains.

However, our previous work did not consider the class imbalance issue, a commonly known obstacle in building a machine learning classifier that can successfully distinguish minority samples from majority ones (Japkowicz and Stephen 2002; Ko et al. 2017). In practice, one would naturally expect the number of malicious web domains to be much fewer than benign ones. Along this line of research, only a limited number of studies (e.g., see Kegelmeyer et al. (2013) and Ye et al. (2010)) had paid attention to data imbalance when developing machine learning models for malicious web domain identification. Generally, resampling techniques such as over-sampling and under-sampling are useful for addressing the data imbalance issue. However, even with advanced sampling techniques, under-sampling may discard some potentially useful data, and over-sampling may increase the likelihood of overfitting (Japkowicz and Stephen 2002). There is no universal agreement about which kinds of methods are better (Zhu et al. 2018).

In this paper, with online credibility and performance data, we propose an integrated resampling approach to address the class imbalance issue in identifying malicious web domains. Specifically, we integrate two types of resampling strategies by starting with over-sampling the minority class moderately, followed by under-sampling the majority class to a similar size as the over-sampled minority class. For over-sampling, we use the Synthetic Minority Over-sampling TEchnique (SMOTE). To under-sample the majority class, an evolutionary under-sampling approach based on the Particle Swarm Optimisation (PSO) algorithm is proposed. Four datasets with different imbalance ratios are used to verify the performance of the proposed integrated resampling approach for malicious web domain identification. Three metrics including *F-measure*, *Geometric Mean* (GMean), and *the area under the ROC curve* (AUC) are considered in the evaluation process. Eight well-known machine learning techniques are included as classifiers to build identification models with the proposed resampling strategy as well as five other commonly used strategies. We apply two-stage statistical tests to check the statistical differences between results obtained by the tested approaches.

The rest of this paper is organised as follows. In Section 2, we review related studies in the areas of malicious site identification and imbalanced data sampling. In Section 3, we present our proposed approach. We then describe our data preparation, evaluation metrics, and experiment settings in Section 4. After that, experimental results are discussed in Section 5. Finally, the conclusion is drawn in Section 6, with some future research directions highlighted.

## 2 Related Work

In this section, we first review related studies on identifying malicious web domains in general, and later discuss studies related to the use of resampling methods for imbalanced data.

### 2.1 Related Studies on Identifying Malicious Attacks

A significant body of knowledge proposing various methods to identify potential phishing websites exists. Several phishing detection techniques that encompass filtering, authentication, phishing report generation and attack tracing have been used in security-related industry implementations. Based on these techniques, various anti-phishing browser toolbars such as TrustWatch (https://www.geotrust.com/comcasttoolbar/), SpoofGuard (https://crypto.stanford.edu/SpoofGuard/) and Netcraft Extension (http://toolbar.netcraft.com/) were developed. According to Tsai et al. (2011), however, these techniques cannot efficiently detect all phishing attacks. They examined 10 anti-phishing tools and discovered that only one of them could consistently detect more than 60% of the phishing websites tested. Their conclusion is echoed by another study from Purkait (2015), who examined the latest commercial browsers and showed that 12 out of 14 tools failed.

By using machine learning-based methods, several studies have explored different types of features in identifying phishing websites (see Sahoo et al. (2017) for a review). Ma et al. built machine learning models based on lexical and host-based features extracted from URLs (Ma et al. 2009; Ma et al. 2009). Xiang et al. proposed Cantina+, a feature-rich machine learning framework that adopts the HTML

Document Object Model and search engine features to detect phishing sites (Zhang et al. 2007; Xiang et al. 2011). Their features include URL, HTML and search engine-based features. An additional feature, a domain top-page similarity feature, was later added to Cantina+ by Sanglerdsinlapachai and Rungsawang (2010). Similar to Cantina+, Whittaker et al. (2010) developed a scalable machine learning classifier to maintain Google's phishing blacklist based on URL and HTML features. In the work of Fu et al. (2006), the Earth Mover's Distance was applied to measure the similarity of web page visual in determining phishing web pages. They converted web page images into low-resolution images to represent the image signatures. They then evaluated similarities of the signatures between web pages. A similar approach was used by Zhang et al. (2011), but they analysed both visual and textual contents and used a Bayesian classifier to detect phishing web pages. In the work of Moghimi and Varjani (2016), four features for evaluating the page resources' identities and four features for identifying the access protocol of page elements were extracted to build a rule-based model for phishing detection in Internet banking. Abutair and Belghith (2017) used both lexical features and some external features dependent on the Jaccard Index and Alexa ranks to build a phishing detection system using a case-based reasoning model. Tan et al. proposed an identity keyword extraction and target domain finder for extracting features from the textual contents of websites and search engines (Tan et al. 2016; Tan et al. 2017). Xiang et al. (2017) built a C5.0-based phishing site detection model by extracting four classes of features, which include address bar-based features, abnormal-based features, JavaScript-based features, and domain-based features. Liu et al. (2018) proposed to combine the character frequency and structural features to detect malicious URLs based on machine learning algorithms. Babagoli et al. (2018) utilised a group of mixed features to build a meta-heuristic nonlinear regression model for detecting phishing websites.

## 2.2 Resampling Techniques for Class Imbalance

Classification of imbalanced data has emerged as one of the most challenging topics in the data mining community (Yang and Wu 2006; Haixiang et al. 2017). It is important not only for malicious site identification, but also for many other real-world classification problems such as credit scoring (Marqués et al. 2013), churn prediction (Zhu et al. 2018), and consumer segmentation (Lo et al. 2015). Existing approaches can be categorised into *data level-based resampling* and *algorithm level* techniques. *Data level-based resampling* techniques resample data before building a model to diminish the skewed class distribution (Chawla et al. 2002; Batista et al. 2004; García and Herrera 2009), and *algorithm level* methods modify existing algorithms or create new ones by taking into account the significance of minority samples (Barandela et al. 2003; Wu and Chang 2005; Zong et al. 2013; Ko et al. 2017). Furthermore, two other groups, which can be designed based on either the data level or algorithm level methods, or both, are also commonly applied in existing studies. They are *cost-sensitive learning* techniques (Xia et al. 2017; Chao and Peng 2018) and *ensemble* methods (Galar et al. 2012; Galar et al. 2013; He et al. 2018). Generally speaking, algorithm level methods are more dependent on the problems, and data level-based resampling techniques are independent of the underlying classifier. The latter can thus be easily applied to different models regardless of whether they are single, cost-sensitive, or ensemble models (Galar et al. 2012). In this study, we focus on the resampling techniques, and some related studies on resampling techniques are reviewed.

There are three kinds of resampling approaches: over-sampling, under-sampling, and hybrid methods. Over-sampling methods increase the size of minority samples by creating new samples or replicating some samples from existing ones. Under-sampling methods eliminate the majority samples to balance the class distribution. Hybrid methods combine both the over- and under-sampling methods. A basic way to implement an over- or under-sampling method is to randomly replicate or remove some samples to balance the data. However, random under-sampling may discard some potentially useful data, and random over-sampling may increase the likelihood of overfitting (Japkowicz and Stephen 2002). To deal with these issues, many advanced methods have been proposed.

Of the advanced over-sampling methods, some typical or recently proposed examples are the SMOTE (Chawla et al. 2002), MWMOTE (majority weighted minority over-sampling) (Barua et al. 2014), graph-based over-sampling (Pérez-Ortiz et al. 2015), diversity-based over-sampling (Bennin et al. 2018), and generative adversarial network-based over-sampling (Douzas and Bacao 2018) techniques. Among them, the SMOTE is the most established and widely used over-sampling method, which creates synthetic minority samples by interpolating existing minority samples along the line segments joining the $k$ minority class nearest neighbours.

Advanced under-sampling methods typically rely on data cleaning techniques. Examples include Wilson's edited nearest neighbour (ENN), one-sided selection, and Tomek Links (Wilson and Martinez 2000; López et al. 2013). The Evolutionary under-sampling method selects the best subset of majority samples by considering the classification performance of a specific model (García and Herrera 2009). Some studies have also proposed combined approaches, such as the integration of SMOTE with ENN or SMOTE with cluster-based under-sampling (Agrawal et al. 2015), or the combination of several over-sampling techniques for generating ensemble models (Huda et al. 2018). It is worth noting that there is no clear agreement about which kind of methods should be followed (Zhu et al. 2018), and sometimes random sampling can also outperform advanced methods (Van Hulse et al. 2007). The choice of a resampling method is therefore data-dependent.

Although data imbalance is inevitable when it comes to malicious site identification, a limited number of studies have paid attention to the class imbalance issue when developing machine learning models in this context. Zhao and Hoi (2013) proposed a cost-sensitive online active learning method to address the imbalance issue for malicious URL detection. Kegelmeyer et al. (2013) considered the use of supervised machine learning models to detect malware executables in network streaming data. They investigated the performance of decision tree ensembles for distinguishing malware and 'goodware' traffic in the presence of concept drift and class imbalance. Ye et al. (2010) proposed a Hierarchical Associative Classifier (HAC) to detect malicious files from a large and imbalanced list. The HAC uses a two-level associative classifier building method to ensure high recall of the minority class before optimising the overall precision.

## 3 The Proposed Approach

This section describes the proposed integrated resampling approach, which is composed of SMOTE-based over-sampling and PSO-based under-sampling.

### 3.1 Synthetic Minority Over-sampling

The SMOTE over-samples the minority class (malicious web domains) by creating synthetic samples along the line segments joining any/all of the $k$ minority class nearest neighbours. Unlike random over-sampling methods, this technique generates synthetic samples rather than replicating minority class samples; therefore, it can overcome the overfitting problem. Briefly, the SMOTE is described in Figure 1 as follows.

---

$O$ is the original dataset
$P$ is the set of minority class instances
**For** each sample $x$ in $P$
    Find the $k$-nearest neighbours (minority class samples) to $x$ in $P$
    Obtain $y$ by randomly selecting one sample from $k$ samples
    $difference = x - y$
    $gap$ = a random number between 0 and 1
    $n = x + difference * gap$
    Add $n$ to $O$
**End For**

---

Figure 1. An overview of the SMOTE (Chawla et al. 2002)

### 3.2 PSO-based Under-sampling

PSO is a population-based meta-heuristic method that simulates social behaviour such as birds flocking to a promising position to achieve precise objectives (e.g., finding food) in a multi-dimensional space by interacting with each other (Kennedy and Eberhart 1997). It possesses advantages such as producing good solutions at a low computational cost and simple implementation. These advantages have led to the widespread use of PSO in many areas (e.g., for parameter optimisation and feature selection) (Cheng et al. 2016; Hu et al. 2017). However, only a limited number of studies have applied PSO to address class imbalance problems. One example can be found in the work of Yang et al. (2009), who applied PSO to resize a training set for medical and biological data mining. Although some studies also applied PSO in models with class imbalance issues, PSO was not used for handling class imbalance itself but for other purposes. For example, Gao et al. (2011) proposed to combine the SMOTE and PSO-based Radial Basis Function (RBF) for two-class imbalanced classification. In their study, the SMOTE was applied to generate synthetic samples for the minority class to balance the data, and PSO was applied to optimise the structure and parameters of the RBF model. Wang et al. (2014) proposed a hybrid classifier by combining the SMOTE with PSO to identify breast cancer, in which PSO was used for feature selection. Furthermore, under-sampling of majority samples is a binary optimisation problem, and PSO has successfully addressed many similar problems (Unler and Murat 2010; Wang et al. 2014; Hafiz et al. 2018). Taking these into consideration, we propose an integrated resampling method by applying a binary version of PSO to under-sample the majority samples and integrating the SMOTE with it.

By applying PSO to under-sample imbalanced data, the algorithm performs its search procedure with a population of individuals called particles. Each particle's status is characterised by its position and velocity. The position of a particle denotes a subset of samples and can be represented as a binary string $X_i = (x_{i1}, x_{i2}, \cdots, x_{ij}, \cdots, x_{in})$, where $n$ is the total number of majority samples (benign web domains). $x_{ij}$ represents a mask of the $j$ th sample in the $i$ th particle. $x_{ij}$ equals 1 if the $j$ th sample is selected, and 0 otherwise. The velocity of a particle denotes the position change from one iteration to the next. By updating their position and velocity iteratively, the particles explore the search space. Figure 2 illustrates the PSO-based resampling approach. As shown in Figure 2, the PSO-based under-sampling process is conducted for benign web domains (majority samples). By combining the malicious web domains (minority samples), particles that denote a subset of benign web domains are evaluated by 5-fold cross validation with a 1-NN (Nearest Neighbour) classifier (Garcá and Herrera 2009). F-measure (see Eq.(2) in Section 4.2) is used as the fitness function. By simulating the social behaviour of birds flocking to a near-optimal position through interactions with each other, PSO updates the position and velocity (see Kennedy and Eberhart (1997) for the updating mechanism of PSO) iteratively, and the optimal subset of benign web domains is explored.

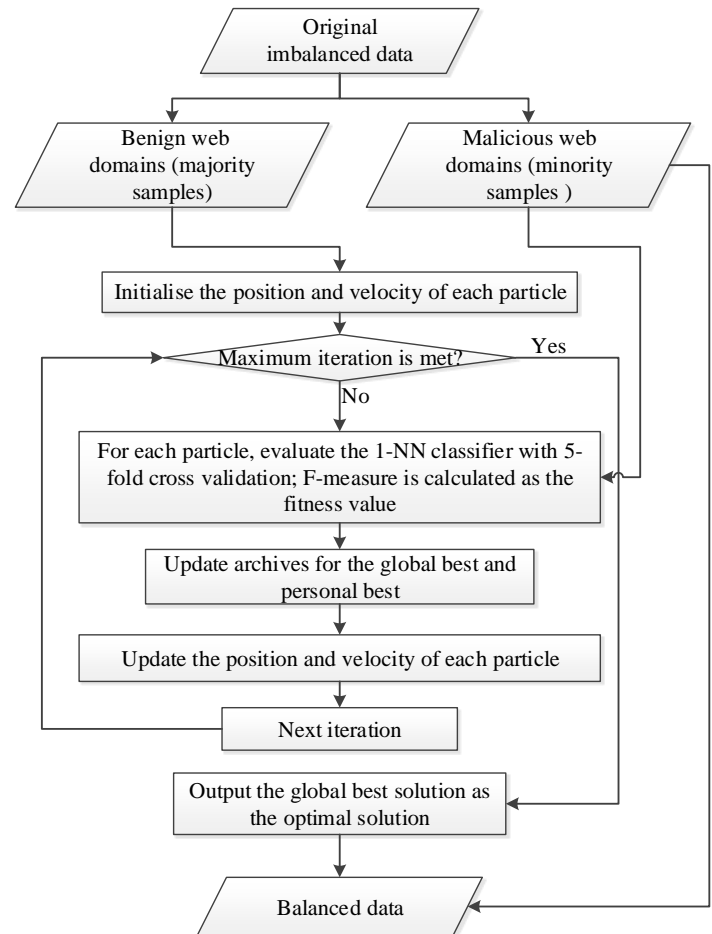

Figure 2. A flowchart showing the PSO-based resampling approach

### 3.3 Integrating the SMOTE and PSO-based Resampling

By integrating the SMOTE and PSO-based resampling, the proposed integrated approach and its evaluation process are illustrated in Figure 3. As can be seen in the figure, the flowchart is divided into two parts: the proposed integrated resampling approach and performance evaluation.

The dashed box in Figure 3 shows the proposed integrated resampling approach. Given an imbalanced dataset, which is a training set in the 10-fold cross validation process in this study, the SMOTE is firstly applied to generate a number of synthetic minority samples in order to increase the number of minority samples. Then, with the resized data obtained by the SMOTE, PSO-based under-sampling is applied to select a subset of majority samples with high identification performance. As suggested by García and Herrera (2009), who showed that a balancing mechanism could help to improve the under-sampling process, the number of selected majority samples is under-sampled by PSO to make it equal to the number of minority samples. Both the SMOTE and evolutionary under-sampling have been successfully applied or highly recommended in previous studies (Chawla et al. 2002; García and Herrera 2009; Wang et al. 2014). PSO has also been successfully used in many binary optimisation problems (Unler and Murat 2010; Wang et al. 2014; Hafiz et al. 2018). Therefore, by integrating SMOTE-based over-sampling and PSO-based under-sampling, the integrated resampling approach is expected to make up their strengths and weaknesses as well as reduce the risk of variability.

Additionally, Figure 3 also depicts the evaluation process. The proposed integrated resampling approach is evaluated by 10-fold cross validation, which is widely applied to evaluate the generalisation performance of an algorithm. The average performance of the cross validation is used for experimental comparison.

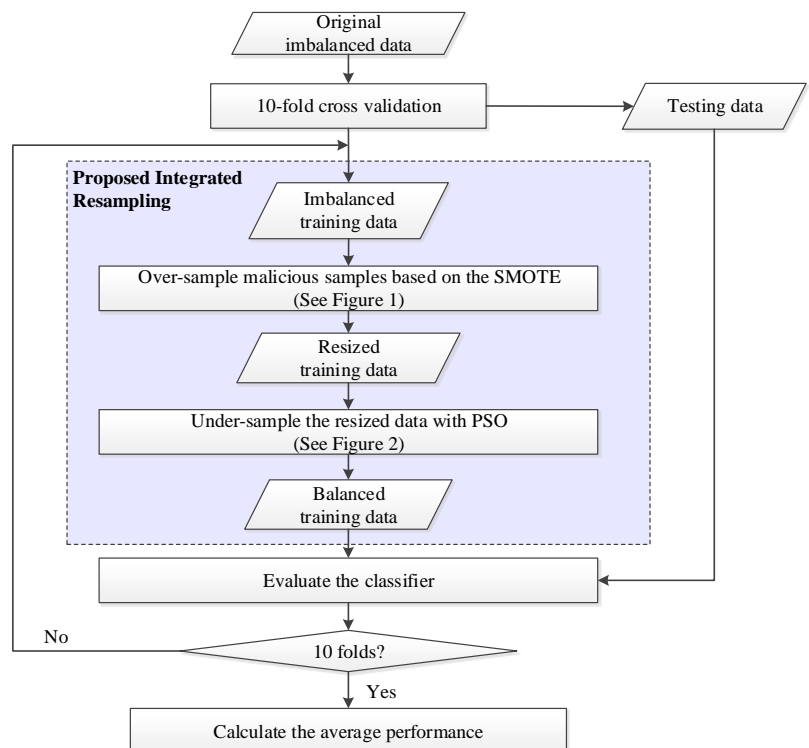

Figure 3. A schematic flowchart of the proposed integrated sampling approach and its evaluation process.

# 4 Experimental Settings

## 4.1 Dataset Preparation

The online credibility and performance data was collected to distinguish malicious web domains from benign ones. Benign web domains were randomly collected from the lists by Alexa (www.alexa.com), while malicious web domains were scanned from up-to-date listings provided by MalwareDomains (www.malwaredomains.com) and PhishTank (www.phishtank.com). These malicious web domains consist of malware propagation domains, attackers' exploit domains, ransomware domains, high-risk domains, and phishing sites. Legitimate web domains that were infected by and known to distribute malware have also been included in the datasets. In total, 22 features were collected from several sources using freely-available API (see Hu et al. (2016) for details). A brief description including sources, names, and descriptions of these features are provided in Table 1. The first feature is the web domain's unique name. The second feature is a label that shows whether a web domain is malicious or benign. The other 20 features are related to performance, social media metrics, and popularity data extracted from Moz, Google, Alexa, LinkedIn, and Facebook.

To examine the effectiveness of our proposed approach in identifying malicious web domains with real-world imbalanced data, we compiled four datasets. Each dataset contains 2000 live Internet domains with the 22 variables/features. Specifically, the datasets are i. D6 with 60% of benign and 40% of malicious web domains, ii. D7 with 70% of benign and 30% of malicious web domains, iii. D8 with 80% of benign and 20% of malicious web domains, and iv. D9 with 90% of benign and 10% of malicious web domains. The ratio, which is defined by the number of benign web domains to the number of malicious web domains, is different for each of the datasets.

Table 1 Data description

| Sources | Features | Descriptions |
|---|---|---|
| MalwareDomains, | DomainName | Name of the web domain. |
| PhishTank, Alexa | isGoodWebsite | Label of a web domain. |
| Moz | MozDomain Authority | Estimated score of how a web domain will perform in search |
|  |  | engine rankings. |
|  | MozPageAuthority | Estimated score of how a page will perform in search engine rankings. |
|  | MozExternalEquityLinks | Number of external links of a web domain. |
|  | MozTotalLinks | Number of links of a web domain. |
|  | MozRank | Moz's global link popularity. |
|  | MozSubdomainRank | Moz's link popularity of subdomain. |
| Google | Google's Page Speed | A metric used to assess the loading speed of a web page. |
|  | NumberResources | Number of HTTP resources loaded by the page. |
|  | NumberHosts | Number of unique hosts referenced by the page. |
|  | NumberStaticResources | Number of static (i.e. cacheable) resources on the page. |
| Alexa | Alexa's rank | A rank used to show the popularity of a web domain regarding the number of visitors. |
|  | Alexa's 1-month reach | The average number of daily unique visitors in the past one month. |
|  | Alexa's 3-month reach | The average number of daily unique visitors in the past three months. |
|  | Alexa's 7-month reach | The average number of daily unique visitors in the past seven months. |
|  | Alexa's median load | The median loading time of a web domain estimated by Alexa. |
| Social media | Facebook shares | Number of shares in Facebook. |
|  | LinkedInCount | Number of shares in LinkedIn. |
|  | LinkedInFCnt | Number of shares in Facebook, provided by LinkedIn. |
|  | LinkedInFCntPlusOne | Number of shares in Google plus, provided by LinkedIn. |
|  | Google plus shares | Number of shares in Google plus. |

## 4.2 Evaluation Metrics

The malicious web domain identification aims to determine if a web domain is benign or malicious. The results of correctly and incorrectly identified samples of each class can be summarised as a confusion matrix, shown in Table 2. Based on Table 2, we could have used *accuracy* (Eq.(1)), which is the percentage of web domains identified correctly over all domains, as a performance measure. However, accuracy cannot distinguish between the numbers of correctly classified samples of each class, especially for the positive class in our malicious domain identification problem with class imbalance. A classifier with very high accuracy may have misclassified the positive classes as negative ones. Accuracy, therefore, is not sufficient to evaluate malicious web domain identification performance with the class imbalance issue.

Table 2 The confusion matrix for malicious web domain identification

| Actual | Predicted | |
|---|---|---|
|  | *Malicious* | *Benign* |
| *Malicious (Positive)* | *TP* (True Positive) | *FN* (False Negative) |
| *Benign (Negative)* | *FP* (False Positive) | *TN* (True Negative) |

$$Accuracy = \frac{TP + TN}{TP + TN + FP + FN} \tag{1}$$

Instead of accuracy, we adopted three metrics that are commonly used for imbalanced classification as follows:

$$F\text{-}measure = \frac{2 * Precision * Recall}{Precision + Recall} \tag{2}$$

$$GMean = \sqrt{TN_{rate} \times TP_{rate}} \tag{3}$$

$$AUC = \frac{1 + TP_{rate} - FP_{rate}}{2} \tag{4}$$

where $TP_{rate} = TP/(TP + FN)$ is the percentage of malicious web domains correctly identified; $TN_{rate} = TN/(FP + TN)$ is the percentage of benign web domains correctly identified;

$FP_{rate} = FP/(FP + TN)$ is the percentage of misclassified benign web domains; $Precision = TP/(TP + FP)$ refers to the percentage of correctly identified malicious web domains over all web domains; and $Recall = TP_{rate}$ evaluates how well a classifier is correctly identifying malicious web domains from all malicious web domains.

F-measure incorporates both precision and recall, and hence it can measure the overall performance of malicious web domain identification models. GMean aims to evaluate the balance of recalls for two classes. If the model is highly biased towards one class, the GMean value will be low. The *AUC* provides a summary of the average performance of an identification model with different parameters. The larger the *AUC* is, the higher the classification potential of the model is.

In this study, we applied F-measure, GMean, and the AUC to evaluate the performances of malicious web domain identification models with different resampling techniques.

### 4.3 Experimental Setup

To examine the performance of the proposed approach, in this study we tested a range of resampling techniques that have been widely used in past studies for comparison purposes. The included techniques and their abbreviations are summarised as follows:

(1) Random Under-sampling (RU), which balances the class distribution by randomly eliminating benign samples.

(2) Random Over-sampling (RO), which randomly replicates malicious samples.

(3) The SMOTE, which is an over-sampling method that creates malicious samples by interpolating several malicious samples that lie together.

(4) PSO-based under-sampling (PSO), where PSO is applied to implement the evolutionary under-sampling approach (see Section 3.2).

(5) The integrated resampling approach (Integrated), which is our proposed approach (see Section 3.3).

(6) Original data without resampling (Ori) as a baseline, that is, the case without any resampling method is also examined.

To verify the performance of our proposed integrated resampling strategy in addressing the class imbalance issue for malicious web domain identification, we applied eight well-known machine learning techniques to build the identification model. Specifically, by following our previous study (Hu et al. 2016), single classifiers such as the artificial Neural Network (NN) (Chiong et al. 2010), Support Vector Machine (SVM) (Vapnik 1995), C4.5 (Quinlan 1993), k-Nearest Neighbour (KNN) (Fukunaga and Narendra 1975) and Naïve Bayes (Bayes and Price 1763), as well as ensemble classifiers such as Random Forest (RF) (Breiman 2001), Adaboost (Freund and Schapire 1997) and Bagging (Breiman 1996), were applied in this study for evaluation purposes. With six resampling approaches and eight classifiers, 48 models were built for each of the four datasets.

All our experiments in this study were carried out in the R programming environment. R packages that have been used to implement the classifiers and the first three resampling strategies are freely available from the CRAN repository (http://cran.r-project.org/). The PSO-based resampling approach and the integrated resampling approach were also implemented in R for this study. Some adjustable parameters in each of the classifiers were tuned by maximising F-measure using 5-fold cross validation. These tuned parameters include: SVM (RBF Kernel, gamma = 0.5, cost = 8), NN (size = 5, maxit = 2000), C4.5 (C = 0.05), KNN (K = 10), RF (ntree = 1000), Adaboost (mfinal = 100, maxdepth = 3), and bagging (mfinal = 100, maxdepth = 5). For RU, RO and the SMOTE, parameters related to the sampling rate were all set based on the imbalance ratio in order to have a balanced dataset. For the PSO-based resampling approach, the swarm size was 40 and the maximum iteration was 2000. Both cognitive and interaction coefficients were set to 2.0.

## 5 Experimental Results and Discussion

### 5.1 Performance Comparison

Tables 3-5 report on the results of different resampling strategies in terms of F-measure, GMean and AUC, respectively. Each table contains the results of different strategies with eight classifiers over the four imbalanced datasets. The best and worst values in each row are highlighted in blue and grey, respectively.

From Tables 3-5, several observations can be made:

(1) The proposed integrated resampling approach performs the best in most cases. The other approaches obtain the best performance only in a few cases. Interestingly, Ori, which builds the malicious domain identification models without any resampling process, produces the best F-measure values in some cases (e.g., C4.5 and bagging-based models on D7, and the NN-based model on D8). In these cases, however, there are no big differences between the Ori-based results and second-best results from the other approaches. On the contrary, the Ori-based model fails to obtain the best performance in terms of GMean and AUC for all cases.

(2) It is worth noting that RU-based models are among the worst in most cases in terms of F-measure, while Ori-based models are the worst in most cases in terms of GMean and AUC with the exception that RU-based models are among the worst for D6. This may be attributed to the fact that RU-based resampling has unconsciously removed many useful samples, especially for D6.

(3) By comparing the performance of different classifiers with different strategies on each dataset, we can observe that the three ensemble models have better performances than the five single models, no matter which resampling strategy or dataset was applied. RF is the best classifier in most cases, and Bayes performs much worse than the others.

By taking the Ori-based model as a baseline, Figures 4-6 illustrate the percentages of improvement for the other five resampling approaches. Labels outside the ring denote the datasets and classifiers. For instance, "D6.SVM" refers to an SVM-based identification model on D6. From these radar charts, we can clearly see that the values of Integrated-based models are at the outermost ring. In other words, our proposed integrated resampling approach is able to obtain much better improvements over Ori-based models compared to the other resampling approaches in most cases. Furthermore, as class imbalance increases, improvements in almost all cases increase too. This indicates that it is beneficial to apply resampling approaches when the data is highly imbalanced.

Table 3. Results of six resampling strategies with eight classifiers over the four datasets (F-measure)

| Data | Classifiers | Ori | RU | RO | SMOTE | PSO | Integrated |
|------|-------------|--------|--------|--------|--------|--------|------------|
| D6 | SVM | 0.8152 | 0.8211 | 0.7976 | 0.8169 | 0.8272 | 0.8370 |
| | NN | 0.8304 | 0.8310 | 0.8256 | 0.8341 | 0.8309 | 0.8439 |
| | KNN | 0.7781 | 0.7781 | 0.7819 | 0.7790 | 0.7727 | 0.7910 |
| | C4.5 | 0.8380 | 0.8375 | 0.7899 | 0.8358 | 0.8444 | 0.8413 |
| | Bayes | 0.6362 | 0.6361 | 0.6422 | 0.6361 | 0.6436 | 0.6459 |
| | RF | 0.8604 | 0.8618 | 0.8997 | 0.8628 | 0.8579 | 0.9200 |
| | adaboost | 0.8491 | 0.8523 | 0.8404 | 0.8529 | 0.8458 | 0.8759 |
| | bagging | 0.8403 | 0.8496 | 0.8450 | 0.8494 | 0.8513 | 0.8522 |
| D7 | SVM | 0.7778 | 0.7766 | 0.7651 | 0.7780 | 0.7702 | 0.7689 |
| | NN | 0.7728 | 0.7711 | 0.7799 | 0.7851 | 0.7833 | 0.8083 |
| | KNN | 0.7109 | 0.7212 | 0.7329 | 0.7190 | 0.7262 | 0.7296 |
| | C4.5 | 0.8054 | 0.7945 | 0.7443 | 0.7983 | 0.8026 | 0.8042 |
| | Bayes | 0.5438 | 0.5437 | 0.5562 | 0.5455 | 0.5542 | 0.5717 |
| | RF | 0.8170 | 0.8220 | 0.8284 | 0.8257 | 0.8381 | 0.8690 |
| | adaboost | 0.8104 | 0.8011 | 0.8062 | 0.8142 | 0.8147 | 0.8542 |
| | bagging | 0.8219 | 0.8097 | 0.8110 | 0.8113 | 0.8150 | 0.8128 |
| D8 | SVM | 0.6653 | 0.6619 | 0.6709 | 0.6839 | 0.6851 | 0.7191 |
| | NN | 0.6949 | 0.6668 | 0.6843 | 0.6929 | 0.6861 | 0.6879 |
| | KNN | 0.6198 | 0.6091 | 0.6107 | 0.6094 | 0.6178 | 0.6455 |
| | C4.5 | 0.7019 | 0.6918 | 0.6595 | 0.6975 | 0.7009 | 0.7035 |
| | Bayes | 0.4050 | 0.4044 | 0.4104 | 0.4047 | 0.4114 | 0.4104 |
| | RF | 0.7387 | 0.7300 | 0.7490 | 0.7683 | 0.7789 | 0.7992 |
| | adaboost | 0.7300 | 0.7069 | 0.7352 | 0.7516 | 0.7470 | 0.7775 |
| | bagging | 0.7046 | 0.7205 | 0.7190 | 0.7252 | 0.7170 | 0.7227 |
| D9 | SVM | 0.5464 | 0.5026 | 0.5322 | 0.5653 | 0.5643 | 0.5910 |
| | NN | 0.5552 | 0.4620 | 0.5258 | 0.5331 | 0.5418 | 0.5802 |
| | KNN | 0.4873 | 0.4297 | 0.4651 | 0.4788 | 0.4862 | 0.5019 |
| | C4.5 | 0.5378 | 0.5309 | 0.5471 | 0.5987 | 0.5505 | 0.5830 |
| | Bayes | 0.2471 | 0.2482 | 0.2608 | 0.2466 | 0.2887 | 0.3427 |
| | RF | 0.6069 | 0.5768 | 0.6675 | 0.6618 | 0.6649 | 0.7151 |
| | adaboost | 0.6185 | 0.5587 | 0.6324 | 0.6453 | 0.6797 | 0.6675 |
| | bagging | 0.5863 | 0.5640 | 0.6121 | 0.6081 | 0.6185 | 0.6531 |

Table 4. Results of six resampling strategies with eight classifiers over the four datasets (GMean)

| Data | Classifiers | Ori | RU | RO | SMOTE | PSO | Integrated |
|------|-------------|--------|--------|--------|--------|--------|------------|
| D6 | SVM | 0.8467 | 0.8321 | 0.8485 | 0.8526 | 0.8445 | 0.8362 |
|    | NN | 0.8598 | 0.8565 | 0.8635 | 0.8610 | 0.8660 | 0.8695 |
|    | KNN | 0.8152 | 0.8211 | 0.8142 | 0.8143 | 0.8178 | 0.8126 |
|    | C4.5 | 0.8664 | 0.8239 | 0.8653 | 0.8664 | 0.8574 | 0.8996 |
|    | Bayes | 0.5074 | 0.5346 | 0.5084 | 0.5080 | 0.5358 | 0.5377 |
|    | RF | 0.8848 | 0.8876 | 0.8873 | 0.9028 | 0.8628 | 0.9266 |
|    | adaboost | 0.8751 | 0.8681 | 0.8791 | 0.8796 | 0.8722 | 0.9032 |
|    | bagging | 0.8675 | 0.8725 | 0.8761 | 0.8766 | 0.8783 | 0.8793 |
| D7 | SVM | 0.8420 | 0.8404 | 0.8497 | 0.8559 | 0.8472 | 0.8459 |
|    | NN | 0.8373 | 0.8562 | 0.8597 | 0.8512 | 0.8578 | 0.8851 |
|    | KNN | 0.7905 | 0.8240 | 0.8154 | 0.8162 | 0.8165 | 0.8204 |
|    | C4.5 | 0.8676 | 0.8298 | 0.8711 | 0.8688 | 0.8680 | 0.8698 |
|    | Bayes | 0.5496 | 0.5819 | 0.5529 | 0.5516 | 0.5798 | 0.5982 |
|    | RF | 0.8673 | 0.8799 | 0.8879 | 0.8910 | 0.8924 | 0.9207 |
|    | adaboost | 0.8626 | 0.8667 | 0.8732 | 0.8737 | 0.8743 | 0.9167 |
|    | bagging | 0.8700 | 0.8744 | 0.8739 | 0.8742 | 0.8670 | 0.8605 |
| D8 | SVM | 0.7767 | 0.8173 | 0.8194 | 0.8289 | 0.8303 | 0.8715 |
|    | NN | 0.7972 | 0.8417 | 0.8425 | 0.8320 | 0.8318 | 0.8341 |
|    | KNN | 0.7454 | 0.7859 | 0.8057 | 0.8033 | 0.8031 | 0.8391 |
|    | C4.5 | 0.8032 | 0.8203 | 0.8427 | 0.8526 | 0.8515 | 0.8546 |
|    | Bayes | 0.5253 | 0.5456 | 0.5242 | 0.5250 | 0.5468 | 0.5455 |
|    | RF | 0.8230 | 0.8730 | 0.8648 | 0.8625 | 0.8961 | 0.9185 |
|    | adaboost | 0.8161 | 0.8489 | 0.8568 | 0.8601 | 0.8548 | 0.8897 |
|    | bagging | 0.7929 | 0.8563 | 0.8633 | 0.8663 | 0.8565 | 0.8990 |
| D9 | SVM | 0.6935 | 0.8140 | 0.7870 | 0.7807 | 0.8060 | 0.8241 |
|    | NN | 0.7034 | 0.8167 | 0.8336 | 0.8335 | 0.8372 | 0.8577 |
|    | KNN | 0.6564 | 0.7988 | 0.8129 | 0.8014 | 0.8276 | 0.8171 |
|    | C4.5 | 0.6991 | 0.8540 | 0.8361 | 0.8375 | 0.8389 | 0.8595 |
|    | Bayes | 0.5788 | 0.5820 | 0.5770 | 0.6151 | 0.6095 | 0.6583 |
|    | RF | 0.7311 | 0.8263 | 0.8490 | 0.8617 | 0.8751 | 0.8961 |
|    | adaboost | 0.7479 | 0.8691 | 0.8402 | 0.8402 | 0.8500 | 0.8618 |
|    | bagging | 0.7222 | 0.8703 | 0.8737 | 0.8737 | 0.8935 | 0.8915 |

Table 5. Results of six resampling strategies with eight classifiers over the four datasets (AUC)

| Data | Classifiers | Ori | RU | RO | SMOTE | PSO | Integrated |
|------|-------------|-----|-----|-----|-------|-----|------------|
| D6 | SVM | 0.8474 | 0.8325 | 0.8490 | 0.8531 | 0.8450 | 0.8367 |
|    | NN | 0.8603 | 0.8569 | 0.8640 | 0.8616 | 0.8665 | 0.8700 |
|    | KNN | 0.8160 | 0.8210 | 0.8151 | 0.8150 | 0.8177 | 0.8125 |
|    | C4.5 | 0.8668 | 0.8251 | 0.8658 | 0.8672 | 0.8582 | 0.9004 |
|    | Bayes | 0.6239 | 0.6353 | 0.6241 | 0.6239 | 0.6367 | 0.6390 |
|    | RF | 0.8852 | 0.8879 | 0.8876 | 0.8895 | 0.8602 | 0.9194 |
|    | adaboost | 0.8756 | 0.8684 | 0.8794 | 0.8800 | 0.8726 | 0.9036 |
|    | bagging | 0.8682 | 0.8729 | 0.8764 | 0.8769 | 0.8786 | 0.8796 |
| D7 | SVM | 0.8442 | 0.8413 | 0.8506 | 0.8565 | 0.8479 | 0.8465 |
|    | NN | 0.8401 | 0.8567 | 0.8604 | 0.8519 | 0.8586 | 0.8590 |
|    | KNN | 0.7954 | 0.8693 | 0.8168 | 0.8373 | 0.8414 | 0.8655 |
|    | C4.5 | 0.8686 | 0.8302 | 0.8716 | 0.8694 | 0.8685 | 0.8703 |
|    | Bayes | 0.6439 | 0.6612 | 0.6462 | 0.6439 | 0.6589 | 0.6798 |
|    | RF | 0.8694 | 0.8883 | 0.8809 | 0.8911 | 0.8996 | 0.9274 |
|    | adaboost | 0.8650 | 0.8677 | 0.8741 | 0.8742 | 0.8748 | 0.8717 |
|    | bagging | 0.8723 | 0.8750 | 0.8744 | 0.8748 | 0.8676 | 0.8611 |
| D8 | SVM | 0.7897 | 0.8198 | 0.8232 | 0.8298 | 0.8312 | 0.8725 |
|    | NN | 0.8081 | 0.8425 | 0.8436 | 0.8330 | 0.8329 | 0.8352 |
|    | KNN | 0.7617 | 0.7879 | 0.8075 | 0.8047 | 0.8050 | 0.8410 |
|    | C4.5 | 0.8135 | 0.8216 | 0.8440 | 0.8535 | 0.8523 | 0.8412 |
|    | Bayes | 0.6341 | 0.6420 | 0.6337 | 0.6332 | 0.6435 | 0.6419 |
|    | RF | 0.8318 | 0.8735 | 0.8678 | 0.8644 | 0.8789 | 0.9054 |
|    | adaboost | 0.8257 | 0.8517 | 0.8597 | 0.8607 | 0.8553 | 0.8903 |
|    | bagging | 0.8065 | 0.8573 | 0.8641 | 0.8669 | 0.8571 | 0.8996 |
| D9 | SVM | 0.7351 | 0.8170 | 0.7995 | 0.7921 | 0.8212 | 0.8545 |
|    | NN | 0.7416 | 0.8187 | 0.8361 | 0.8464 | 0.8464 | 0.8604 |
|    | KNN | 0.7069 | 0.8005 | 0.8148 | 0.8035 | 0.8174 | 0.8311 |
|    | C4.5 | 0.7388 | 0.8557 | 0.8416 | 0.8397 | 0.8415 | 0.8512 |
|    | Bayes | 0.6608 | 0.6627 | 0.6599 | 0.6844 | 0.6904 | 0.7467 |
|    | RF | 0.7636 | 0.8364 | 0.8401 | 0.8663 | 0.8701 | 0.9094 |
|    | adaboost | 0.7708 | 0.8703 | 0.8464 | 0.8464 | 0.8683 | 0.8717 |
|    | bagging | 0.7578 | 0.8714 | 0.8755 | 0.8755 | 0.8886 | 0.8948 |

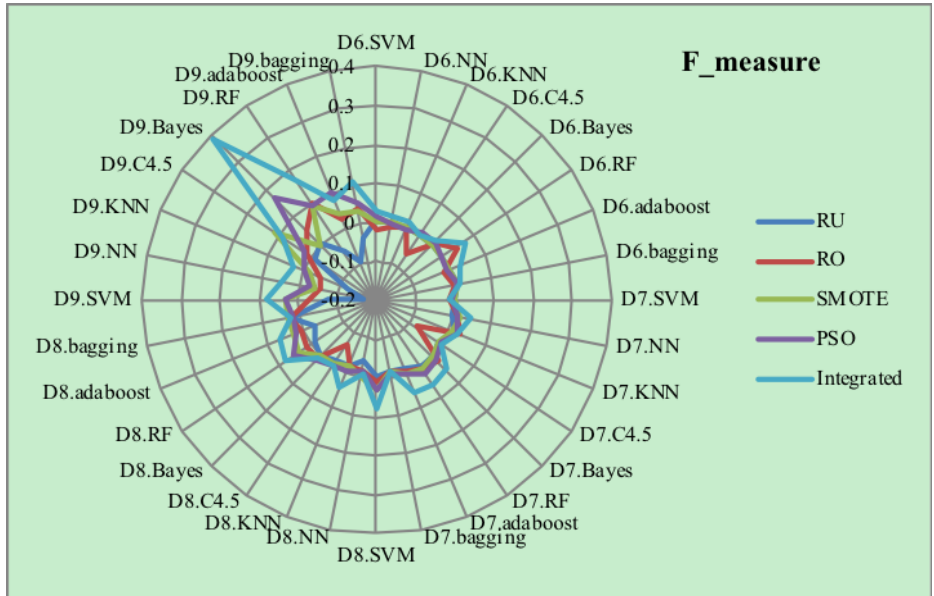

Figure 4. Percentages of improvement taking the Ori-based model as a baseline (F_measure)

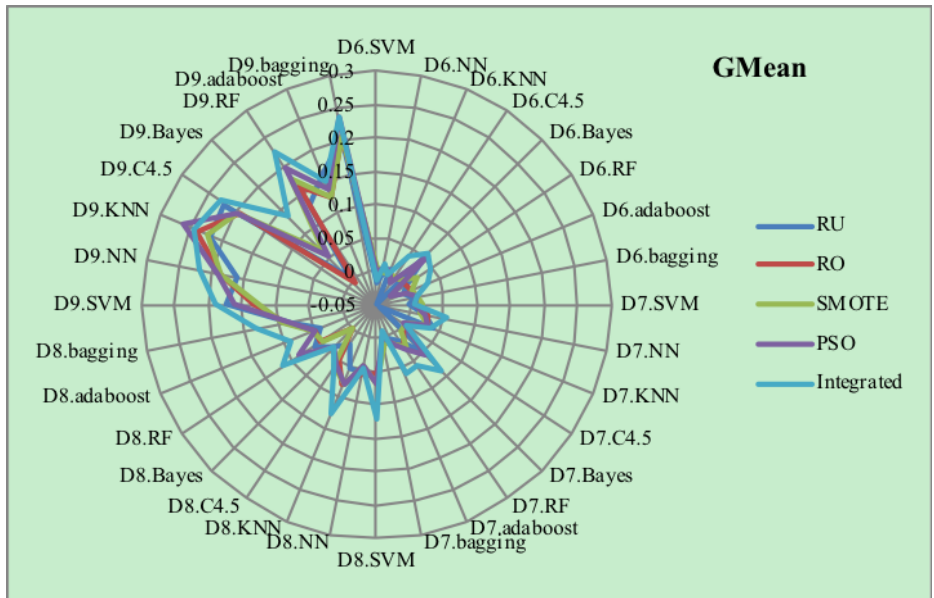

Figure 5. Percentages of improvement taking the Ori-based model as a baseline (GMean)

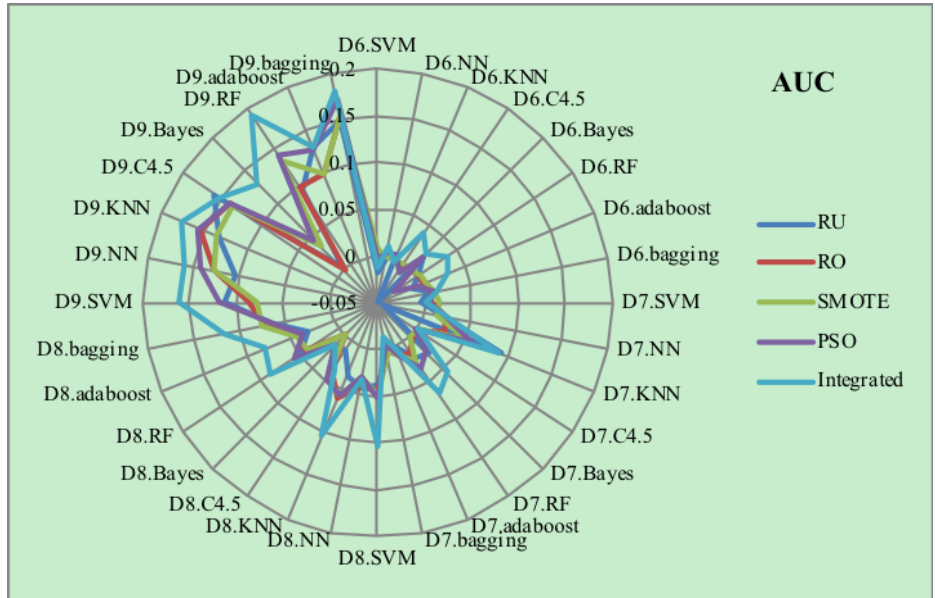

Figure 6. Percentages of improvement taking the Ori-based model as a baseline (AUC)

## 5.2 Statistical Analysis

To verify the statistical significance of the results obtained by different resampling approaches on all datasets, we applied a two-stage procedure for statistical analysis (Demšar 2006; Weise and Chiong 2015). We first used the Friedman and Iman-Davenport tests to check the null hypothesis that there is no significant difference between results obtained by all the models based on their measured average rankings in the four datasets with different classifiers. After confirming that the null hypothesis was rejected, we used three post-hoc tests, namely the Bonferroni-Dunn, Holm, and Hochberg tests, to detect statistical differences between the control approach (one with the lowest Friedman rank) and other approaches in comparison. Tables 6 and 7 show the statistical test results of this two-stage procedure in terms of F-measure, GMean, and AUC.

In Table 6, we can see the results of Friedman and Iman-Davenport tests for all resampling approaches over all datasets and classifiers. Rank_F_m, Rank_GM, and Rank_AUC are the average ranks of the six resampling approaches from Tables 3-5, respectively. The higher the rank is, the better the approach performs. From the table, we observe that RU has the lowest rank in terms of F-measure, and Ori has the lowest rank in terms of GMean and AUC, and the proposed integrated approach has the highest rank for all metrics. That is, with eight classifiers over four datasets, the proposed integrated resampling approach performs best according to three different evaluation metrics. More importantly, the Friedman and Iman-Davenport test values are all significant at the 0.001 level, indicating that significant differences exist among the results. Given the significant differences, three post-hoc tests were conducted. The best performing model based on each metric was selected as the control method for pairwise comparisons with other models in the post-hoc tests.

Table 6. Results of the Friedman and Iman-Davenport tests for different resampling approaches

| Approach | Rank_F_m*# | Rank_GM*# | Rank_AUC*# |
|---|---|---|---|
| Ori | 4.0469 | 5.3438 | 5.3125 |
| RU | 5.0312 | 3.9688 | 3.9375 |
| RO | 4.1406 | 3.5625 | 3.3750 |
| SMOTE | 3.3906 | 3.3125 | 3.4062 |
| PSO | 2.9062 | 3.0000 | 3.0000 |
| Integrated | 1.4844 | 1.8125 | 1.9688 |

\* Significant at the 0.001 level with the Friedman test;
\# Significant at the 0.001 level with the Iman-Davenport test.

Table 7 shows the adjusted *p*-values for each comparison in terms of the three metrics. Considering the levels of significance $\alpha$ =0.05 and $\alpha$ =0.1, *p*-values that are less than 0.05 and 0.1 have been marked in *italic* and **bold** respectively. The proposed integrated resampling approach has been selected as the control algorithm because of its highest rank. According to the results of the three post-hoc tests, pairwise comparisons between Integrated and other models show significant differences, with an exception that the difference between Integrated and PSO is not significant in terms of AUC based on the Bonferroni-Dunn test (pBonf equals 1.4E-01). In summary, the proposed integrated approach has performed significantly better than the other approaches in almost all cases.

Table 7. Comparison results of the post-hoc tests for different resampling approaches

| Integrated vs. Others | F-measure | | | GMean | | | AUC | | |
|---|---|---|---|---|---|---|---|---|---|
| | pBonf | pHolm | pHochberg | pBonf | pHolm | pHochberg | pBonf | pHolm | pHochberg |
| Ori | *0* | *0* | *0* | *0* | *0* | *0* | *0* | *0* | *0* |
| RU | *0* | *0* | *0* | *2.0E-05* | *1.6E-05* | *1.6E-05* | *1.3E-04* | *1.0E-04* | *1.0E-04* |
| SMOTE | *0* | *0* | *0* | *9.1E-04* | *5.5E-04* | *5.5E-04* | *1.1E-02* | *6.3E-03* | *5.3E-03* |
| RO | *2.3E-04* | *9.2E-05* | *9.2E-05* | *6.7E-03* | *2.7E-03* | *2.7E-03* | *1.3E-02* | *6.3E-03* | *5.3E-03* |
| PSO | *1.2E-02* | *2.4E-03* | *2.4E-03* | **5.6E-02** | *1.1E-02* | *1.1E-02* | 1.4E-01 | *2.7E-02* | *2.7E-02* |

# 6 Conclusion, Implications, and Future Research Directions

Machine learning models have performed well in identifying malicious web domains on balanced datasets with online credibility and performance data in our previous study. By focusing on the imbalanced class distribution problem, which is a common obstacle in malicious identification modelling, we proposed an integrated resampling approach to address the class imbalance issue and improve the performance of malicious web domain identification. In our proposed approach, the SMOTE is first used to over-sample malicious web domains; then a PSO-based under-sampling technique is applied with the aim of selecting the optimal subset of benign samples and getting a balanced dataset. With eight well-known machine learning methods, we built malicious web domain identification models with different resampling approaches and evaluated their performances in terms of F-measure, GMean and AUC. Comprehensive experiments using six resampling approaches with different classifiers over four datasets showed that the proposed integrated approach performs significantly better than other approaches in most cases. The experimental results were validated via two-stage statistical tests.

Our work contributes to the existing literature in several ways. First, with online credibility and performance data applied for identifying malicious web domains, we addressed the class imbalance issue in this study. As discussed in Sections 1 and 2, limited previous studies had considered class imbalance when developing models for malicious web domain identification. Second, we proposed an integrated resampling approach, by using the SMOTE for over-sampling and PSO for under-sampling. Although some hybrid resampling approaches were presented in previous studies (Agrawal et al. 2015; Huda et al. 2018), integrating SMOTE-based over-sampling and PSO-based under-sampling has not been explored before. Third, by compiling real-world datasets with different imbalance ratios and testing them with eight machine learning methods, the proposed integrated resampling approach was comprehensively evaluated. Our experimental results provided a clear indication that the proposed integrated approach can be a promising alternative for detecting malicious web domains in real life.

From a practical point of view, this work is expected to set the path for a safer online environment for general Internet users through the development of an adaptive machine learning approach with suitable resampling strategies that can effectively prevent online malicious attacks. The adaptive machine learning approach can automate the identification of malicious links and websites, thereby preventing Internet users from unknowingly sharing their private or confidential information. In addition, it can reduce the risk of potential malicious attacks that normally happen in a stealth mode by providing warnings to the users when they receive or visit a malicious online link. The use of machine learning and real-time online data further reduces the need to frequently update the signatures of anti-virus or anti-malware software. This means Internet users do not need to have in-depth security knowledge or follow regular reminders to keep their online platforms safe.

This study has a few limitations that should be addressed in the near future. First, only four imbalanced web domain datasets and five well-known resampling techniques were applied for evaluation purposes. For future work, we would like to carry out more comparative analysis with other well-designed resampling approaches and evaluate these approaches on more imbalanced web domain datasets. Second, we only utilised the online credibility and performance data in building malicious web domain detection models. A future study could extend and refine the identification model by integrating our collected credibility and performance data with other kinds of features, such as lexical and host-based features. Third, although resampling is an effective way to address the class imbalance issue in detecting malicious web domains, it is also worthwhile to design novel cost-sensitive or ensemble models to address the imbalanced class distribution issue and to improve the performance of malicious site identification.

# Acknowledgements

This work was supported by the Natural Science Foundation of China (Grants 71571080 and 71601147), the Fundamental Research Funds for the Central Universities (104-413000017), and the China Postdoctoral Science Foundation (2015M582280). We would like to thank the handling editor and two anonymous reviewers for their valuable comments and suggestions on the previous version of this paper.